\documentclass[sigconf]{acmart}

\usepackage{multirow}
\usepackage{listings}
\usepackage[capitalize]{cleveref}

\allowdisplaybreaks

\definecolor{codegreen}{rgb}{0,0.6,0}
\definecolor{codegray}{rgb}{0.5,0.5,0.5}
\definecolor{codepurple}{rgb}{0.58,0,0.82}
\definecolor{backcolour}{rgb}{0.95,0.95,0.92}

\lstdefinestyle{mystyle}{
    language=Python,
    backgroundcolor=\color{backcolour},   
    commentstyle=\color{codepurple}\bfseries,
    keywordstyle=\color{codegreen}\bfseries,
    numberstyle=\tiny\color{codegray},
    stringstyle=\color{red},
    basicstyle=\ttfamily\footnotesize, 
    breakatwhitespace=false,         
    breaklines=true,                 
    captionpos=b,                    
    keepspaces=false,                 
    numbers=right,                    
    numbersep=1pt,                  
    showspaces=false,                
    showstringspaces=false,
    showtabs=false,                  
    tabsize=2,
    morekeywords={yield},
    emph ={problem_op, mixing_op},
    emphstyle=\color{blue},
    deletekeywords=[2]{compile},
}

\newcommand{\ie}{\textit{i}.\textit{e}.}

\crefname{section}{Sec.}{Secs.}
\Crefname{section}{Section}{Sections}
\Crefname{table}{Table}{Tables}
\crefname{table}{Tab.}{Tabs.}

\AtBeginDocument{%
  }

\copyrightyear{2023}
\acmYear{2023}
\setcopyright{licensedothergov}
\acmConference[CIKM '23]{Proceedings of the 32nd ACM International Conference on Information and Knowledge Management}{October 21--25, 2023}{Birmingham, United Kingdom}
\acmBooktitle{Proceedings of the 32nd ACM International Conference on Information and Knowledge Management (CIKM '23), October 21--25, 2023, Birmingham, United Kingdom}
\acmPrice{15.00}
\acmDOI{10.1145/3583780.3615006}
\acmISBN{979-8-4007-0124-5/23/10}

\begin{document}

\title[PCT-CycleGAN]{PCT-CycleGAN: Paired Complementary Temporal Cycle-Consistent Adversarial Networks for Radar-Based Precipitation Nowcasting}

\author{Jaeho Choi}
\authornote{
Corresponding author.
This research was supported by the ``Development of radar based severe weather nowcasting technology (KMA2021-03122)'' of ``Development of integrated application technology for Korea weather radar'' project funded by the Weather Radar Center, Korea Meteorological Administration.
}
\affiliation{%
  \institution{Korea Meteorological Administration}
  \city{Seoul}
  \country{Republic of Korea}
}
\email{jaehochoi2021@korea.kr}

\author{Yura Kim}
\affiliation{%
  \institution{Korea Meteorological Administration}
  \city{Seoul}
  \country{Republic of Korea}
}
\email{yrkim110@korea.kr}

\author{Kwang-Ho Kim}
\affiliation{%
  \institution{Korea Meteorological Administration}
  \city{Seoul}
  \country{Republic of Korea}
}
\email{khkim777@korea.kr}

\author{Sung-Hwa Jung}
\affiliation{%
  \institution{Korea Meteorological Administration}
  \city{Seoul}
  \country{Republic of Korea}
}
\email{shjung95@korea.kr}

\author{Ikhyun Cho}
\affiliation{%
  \institution{Korea Meteorological Administration}
  \city{Seoul}
  \country{Republic of Korea}
}
\email{ehcho@kma.go.kr}

\renewcommand{\shortauthors}{Jaeho Choi, Yura Kim, Kwang-Ho Kim, Sung-Hwa Jung, and Ikhyun Cho}

\begin{abstract}
The precipitation nowcasting methods have been elaborated over the centuries because rain has a crucial impact on human life.
Not only quantitative precipitation forecast (QPF) models and convolutional long short-term memory (ConvLSTM), but also various sophisticated methods such as the latest MetNet-2 are emerging.
In this paper, we propose a paired complementary temporal cycle-consistent adversarial networks (PCT-CycleGAN) for radar-based precipitation nowcasting, inspired by cycle-consistent adversarial networks (CycleGAN), which shows strong performance in image-to-image translation.
PCT-CycleGAN generates temporal causality using two generator networks with forward and backward temporal dynamics in paired complementary cycles.
Each generator network learns a huge number of one-to-one mappings about time-dependent radar-based precipitation data to approximate a mapping function representing the temporal dynamics in each direction.
To create robust temporal causality between paired complementary cycles, novel connection loss is proposed.
And torrential loss to cover exceptional heavy rain events is also proposed.
The generator network learning forward temporal dynamics in PCT-CycleGAN generates radar-based precipitation data 10 minutes from the current time.
Also, it provides a reliable prediction of up to 2 hours with iterative forecasting.
The superiority of PCT-CycleGAN is demonstrated through qualitative and quantitative comparisons with several previous methods.
\end{abstract}

\ccsdesc[500]{Computing methodologies~Computer vision}
\ccsdesc[500]{Applied computing~Earth and atmospheric sciences}
\ccsdesc[500]{Applied computing~Forecasting}
\ccsdesc[500]{Computing methodologies~Machine learning}

\keywords{PCT-CycleGAN; generative model; weather radar; precipitation nowcasting}

\maketitle 

\section{Introduction}
\label{sec:intro}
Rain has had a close influence on human life since ancient times.
Therefore, precipitation forecasting has always been a topic of human interest.
There have been many attempts to predict precipitation from a meteorological point of view.
Typically, the development of numerical weather prediction (NWP) models has been at the center of these attempts~\cite{sun2014, wilson1998}.
The NWP models support good quantitative precipitation forecasting across a wide timeline.
However, the NWP models cannot guarantee accuracy in the first few hours because they develop the convective-scale structures in the early stages of forecasting~\cite{short2022}.
This limitation is called a spin-up problem.
To solve this problem, the importance of nowcasting, which focuses on short-term forecasting, has increased.

In meteorology, nowcasting generally refers to a model or forecasting that accurately predicts weather within about two hours from now.
In particular, the precipitation nowcasting is generally based on weather radar observations.
Although various quantitative precipitation forecast (QPF) models exist~\cite{georgakakos1984, turner2004}, perfect precipitation nowcasting is always a challenge.
With the great success of generative models such as generative adversarial networks (GANs)~\cite{goodfellow2014, goodfellow2020}, there are increasing attempts to use these for precipitation nowcasting.
However, there are very few successful cases.
This is because precipitation data include various distributions that are very difficult to learn, so in-depth analysis and understanding of precipitation data through collaboration between artificial intelligence researchers and meteorologists are required.

\begin{figure}[t]
    \centering
    \includegraphics[width=0.88\columnwidth]{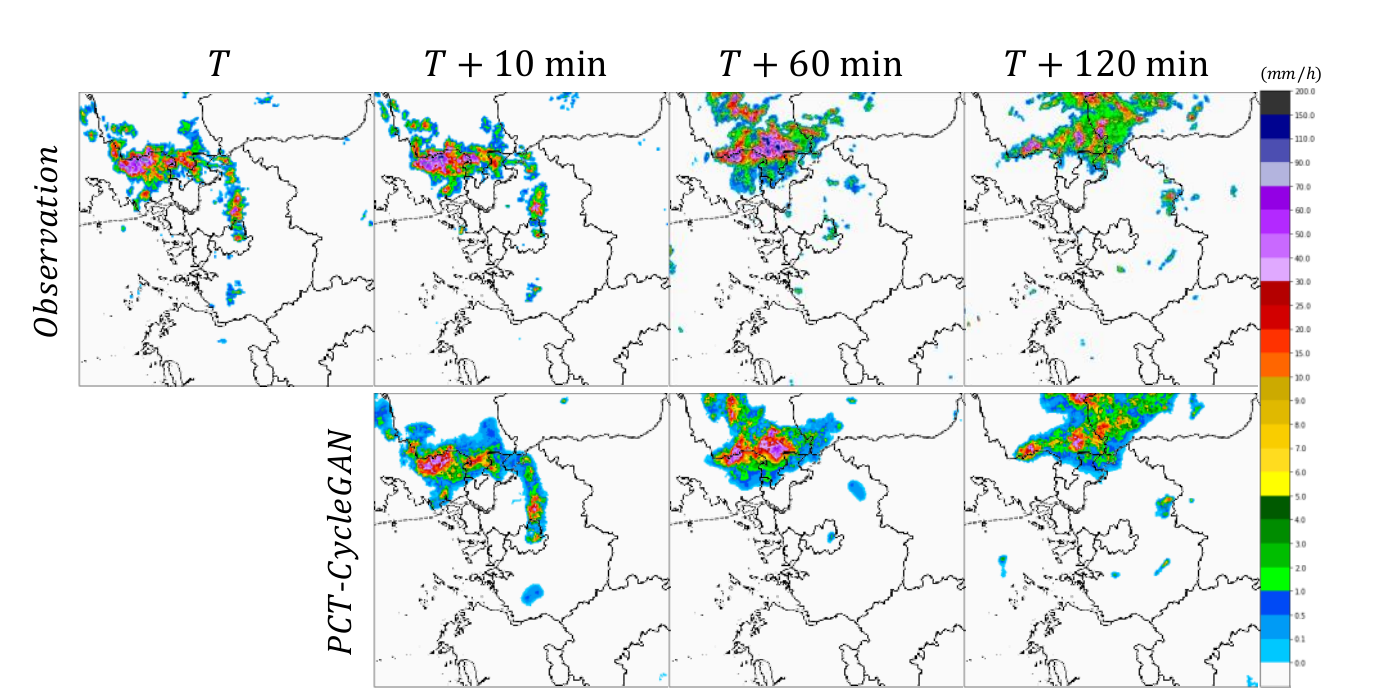}
    \vspace{-4mm}
    \caption{
    Precipitation nowcasting in Seoul using paired complementary temporal cycle-consistent adversarial networks (PCT-CycleGAN).
    The starting point \textit{T} is 2022-08-07 00:00 (UTC). 
    }
    \vspace{-5mm}
    \label{fig:teaser}
\end{figure}
Inspired by cycle-consistent adversarial networks (CycleGAN)~\cite{zhu2017}, we propose a novel precipitation nowcasting method which is called paired complementary temporal cycle-consistent adversarial networks (PCT-CycleGAN) as shown in~\Cref{fig:teaser}.
PCT-CycleGAN trains two generators and two discriminators in paired complementary cycles.
One generator learns mapping into one step future (forward temporal dynamics), and the other learns mapping into one step past (backward temporal dynamics).
The two discriminators determine whether the mapping into the future and the past are appropriate, respectively.
Here, we focus on a regional part of the mesoscale convective system.
Therefore, the mapping from the distributions of current precipitation echoes to the distributions of future precipitation echoes and vice versa can be finitely modeled.
In addition, the mapping should guarantee the temporal causality that does not reverse or stop.
PCT-CycleGAN removes the identity loss added in the vanilla CycleGAN and adds paired data and connection loss concepts to guarantee robust temporal causality.
We evaluate PCT-CycleGAN on three different datasets according to the K\"{o}ppen-Geiger climate classification criteria~\cite{beck2018}.
As a result, PCT-CycleGAN outperforms the existing representative QPF model, the McGill algorithm for precipitation nowcasting by lagrangian extrapolation (MAPLE)~\cite{turner2004}, during a lead time of two hours.
It also outperforms convolutional long short-term memory (ConvLSTM)~\cite{shi2015}, a representative recurrent neural networks (RNNs) series model for precipitation nowcasting.
And comparison with MetNet-2~\cite{espeholt2022}, the latest deep learning model for precipitation nowcasting, is also performed.
The critical success index (CSI)~\cite{schaefer1990}, the peak signal-to-noise ratio (PSNR), and the structural similarity index measure (SSIM) are used for comparative evaluation~\cite{wang2004}.

The remainder of this paper is structured as follows.
\Cref{sec:related}~reviews related work briefly.
\Cref{sec:proposed}~describes the details of PCT-CycleGAN such as proposed objective functions and network architectures.
\Cref{sec:results}~demonstrates the superiority of our proposed method through evaluation and discusses its limitations.
Finally, we conclude the paper in~\Cref{sec:conclusion}.

\section{Related Work}
\label{sec:related}
$\bullet$~{\bf Generative Adversarial Networks.}
GANs are one of the most innovative ideas that have succeeded in the field of artificial intelligence~\cite{goodfellow2014, li2019}.
GANs consist of two kinds of networks, which are called generator and discriminator.
The goal of the generator is to learn the probability distribution that generated the training data~\cite{goodfellow2020}.
In other words, the generator becomes possible to generate fake data that is difficult to discriminate from real data through the training process.
The goal of the discriminator is to distinguish between real and fake data.
During the training process, the quality of generated fake data is increased via the discriminator's feedback.

\vspace{0.5mm}
\noindent
$\bullet$~{\bf Least Squares Generative Adversarial Networks.}
The key to the optimization of the regular GANs is to minimize the Jensen-Shannon divergence (JSD) between the model's distribution and the data-generating process~\cite{goodfellow2014}.
However, this approach causes weaknesses such as mode collapse and learning instability.
Thus, there were attempts to improve the objective function of regular GANs.
Instead of the minimum of JSD, Wasserstein GANs (WGANs)~\cite{arjovsky2017}, which find the minimum value of earth mover's distance (EMD), and its improved version with gradient penalty (GP), WGAN-GP appeared~\cite{gulrajani2017}.
And least squares GANs (LSGANs)~\cite{mao2017, mao2019}, which find the minimum value of Pearson ${\chi}^2$ divergence, also appeared.
In particular, LSGANs are used in many GAN-based applications because of their simple implementation and good performance~\cite{chen2021, dewi2021, lee2022, mukherkjee2022, zou2019}.
We adopt the adversarial losses of LSGANs as a part of our optimization functions because of their faster and more stable convergence rate.

\vspace{0.5mm}
\noindent
$\bullet$~{\bf Cycle-Consistent Adversarial Networks.}
The image-to-image translation is one of the popular topics in the deep learning area, and various studies have been made~\cite{gatys2016, huang2018, isola2017, ko2022, liu2017, liu2016, park2020, zhu2017}.
In particular, techniques based on GANs become a golden key in the field of image-to-image translation because of their powerful ability to formulate and generate high-resolution images~\cite{zhang2022}.
As a pioneer, Pix2Pix succeeded in paired image-to-image translation using conditional GANs~\cite{isola2017}.
Based on this, CycleGAN solved the unpaired image-to-image translation using cycle-consistency loss~\cite{zhu2017}.
CycleGAN, which assumes the relationship between the two domains as a bijection, is shown good performance in various fields~\cite{jiang2022, kwon2019, li2021, mathew2020, park2020}.
In particular, the applied study of Kwon and Park used CycleGAN to predict the next frame for the video~\cite{kwon2019}.
Their proposed multi-input-single-output generator, which should maintain the input sequence of images, does not create temporal causality via CycleGAN by itself.
However, PCT-CycleGAN obtains temporal causality from its own losses without forced input sequences.

\vspace{0.5mm}
\noindent
$\bullet$~{\bf Precipitation Nowcasting via Deep Learning.}
Accurate precipitation nowcasting is a long-standing challenge in meteorology because rain has a huge impact on human life.
Various meteorological models for precipitation nowcasting were built~\cite{fox2005, germann2002, germann2004, metta2009, pierce2004, sun2014, turner2004}, but it is always a difficult problem for humans to accurately consider complex factors for natural phenomena.
As the deep learning model has great success in various fields, several attempts are active to apply this to precipitation nowcasting.
The ConvLSTM, which can make better predictions than the method based on the optical flow of consecutive radar maps, appeared in $2015$~\cite{shi2015}.
Following the success of ConvLSTM, various models for precipitation nowcasting based on RNNs emerged to capture temporal changes in radar echoes~\cite{asanjan2018, chen2020, chen2022, jose2022, luo2021, shi2017}.
In $2020$, RainNet, which uses an encoder-decoder architecture such as U-Net, appeared~\cite{ayzel2020}.
After the birth of RainNet, many models for precipitation nowcasting used U-Net architecture~\cite{badrinarayanan2017, han2022, ronneberger2015, trebing2021}.
In recent years, ResNet-based models are in the spotlight~\cite{he2016}.
Two representative models using residual blocks of ResNet, the deep generative model of rainfall (DGMR) and MetNet-$2$ showed successful results for several hours of precipitation forecasting~\cite{espeholt2022, ravuri2021}.
To obtain temporal causality, DGMR and MetNet-$2$ use the convolutional gated recurrent unit (ConvGRU) and ConvLSTM, respectively~\cite{siam2017}.
In other words, both models also depend on RNNs.
We adopt residual blocks for performance, but not RNNs.
Nevertheless, PCT-CycleGAN guarantees great predictive results within the scope of nowcasting.

\begin{figure*}[t]
    \centering
    \includegraphics[width=1\textwidth]{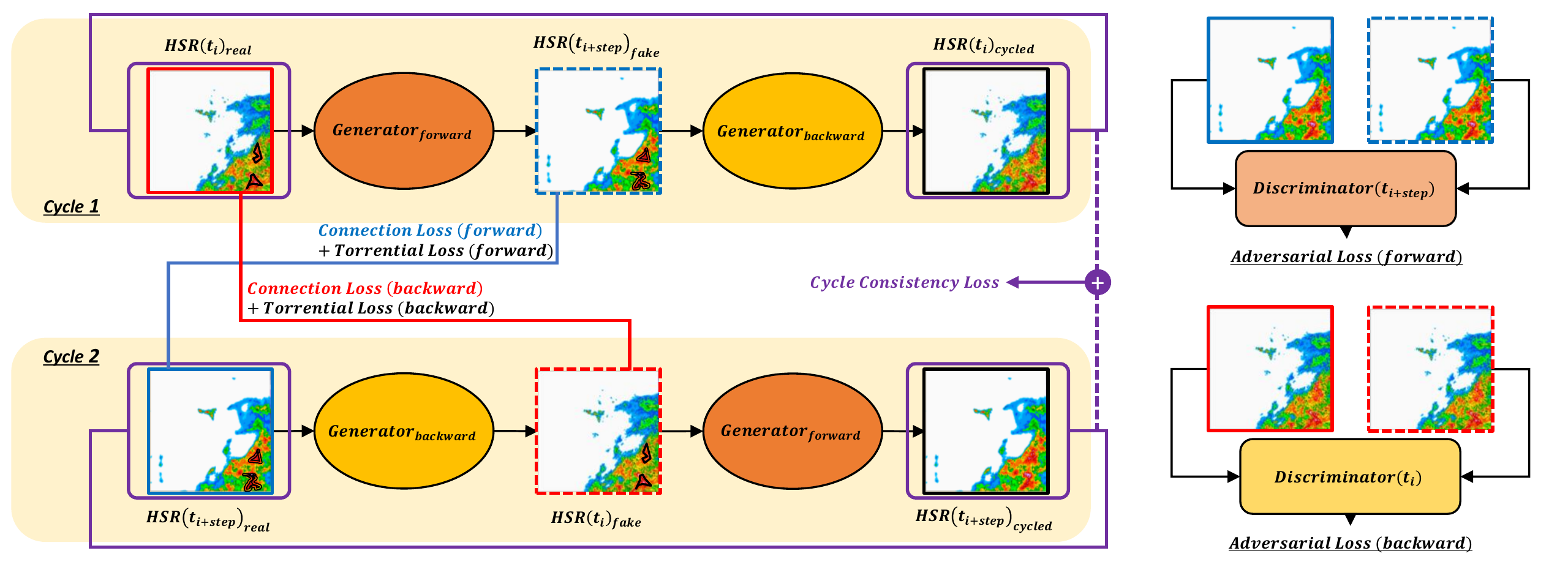}
    \vspace{-7mm}
    \caption{
    Training schematic of PCT-CycleGAN for precipitation nowcasting with a batch size of 1.
    For the clear and concise explanation, $\textit{Generator}_{forward}$, $\textit{Generator}_{backward}$, $\textit{Discriminator}(t_{i+step})$, and $\textit{Discriminator}(t_{i})$ are denoted as $\textit{G}_f$, $\textit{G}_b$, $\textit{D}(t_{i+step})$, and $\textit{D}(t_{i})$, respectively.
    Two generators ($\textit{G}_f$ and $\textit{G}_b$) and two discriminators ($\textit{D}(t_{i+step})$ and $\textit{D}(t_{i})$) are trained via two cycles.
    $\textit{G}_f$ and $\textit{G}_b$ generate the fake composite hybrid surface rainfall (HSR) of the next time step and the previous time step from the input HSR, respectively.
    $\textit{D}(t_{i+step})$ discriminates the fake HSR generated by $\textit{G}_f$ and the real HSR.
    Similarly, $\textit{D}(t_{i})$ discriminates the fake HSR generated by $\textit{G}_b$ and the real HSR.
    $\textit{Cycle}$~1 generates the one-step fake future from the real $\textit{HSR}(t_i)$, and generates cycled $\textit{HSR}(t_i)$ again from the generated future,~\ie, $\textit{Cycle}~\text{1} \triangleq\textit{HSR}(t_i)_{real}\rightarrow \textit{HSR}(t_{i+step})_{fake}\rightarrow \textit{HSR}(t_i)_{cycled}$.
    $\textit{Cycle}$~2 generates the one-step fake past from the real $\textit{HSR}(t_{i+step})$, and generates cycled $\textit{HSR}(t_{i+step})$ again from the generated past,~\ie, $\textit{Cycle}~\text{2}\triangleq\textit{HSR}(t_{i+step})_{real}\rightarrow \textit{HSR}(t_{i})_{fake}\rightarrow \textit{HSR}(t_{i+step})_{cycled}$.
    The final goal of the model is to make real, fake, and cycled HSR identical.
    In other words, it should be $\textit{HSR}(t_i)_{real}\approx\textit{HSR}(t_i)_{fake}\approx\textit{HSR}(t_i)_{cycled}$ and $\textit{HSR}(t_{i+step})_{real}\approx\textit{HSR}(t_{i+step})_{fake}\approx\textit{HSR}(t_{i+step})_{cycled}$ after training.
    Black marks in $\textit{HSR}(t_i)_{real}$, $\textit{HSR}(t_i)_{fake}$, $\textit{HSR}(t_{i+step})_{real}$, and $\textit{HSR}(t_{i+step})_{fake}$ represent precipitation above the threshold used for the computation of torrential losses.
    }
    \vspace{-4mm}
    \label{fig:schematic}
\end{figure*}
\section{Proposed Method}
\label{sec:proposed}
As described in~\Cref{fig:schematic}, PCT-CycleGAN consists of two cycles and uses composite hybrid surface rainfall (HSR)~\cite{oh2018}, the radar-based precipitation data.
A total of four different networks are trained in two cycles.
$\textit{Generator}_{forward}$ and $\textit{Generator}_{backward}$ learn the mapping from present to future and from future to present, respectively.
$\textit{Discriminator}(t_{i+step})$ and $\textit{Discriminator}(t_i)$ evaluate the mapping accuracy of $\textit{Generator}_{forward}$ and $\textit{Generator}_{backward}$, respectively.

For clarity, we briefly explain the notations used in the rest of the paper.
A sample of data, $\textit{HSR}_i$, used as input to the model is denoted as follows:
\begin{align}
    \textit{HSR}_{i}\triangleq\Big(\textit{HSR}(t_i)_{real},~\textit{HSR}(t_{i+step})_{real}\Big),
    \label{eq:hsr_data_definition}
\end{align}
where $t_i$ is an arbitrary element of $T_n=\{t_1,t_2,\cdots,t_n\}$, the arithmetic sequence of discrete-time; $n\in\mathbb{Z}^{+}$, $i\in\mathbb{Z}^{+}$, $step\in\mathbb{Z}^{+}$, and ${i+step}\leq n$.
Note that $\textit{HSR}(t_i)_{real}$ represents real HSR data at time $t_i$.
$\textit{Cycle}~1$, which predicts the future from the known present and predicts the present again from the predicted future, is defined as follows:
\begin{align}
    \textit{Cycle}~1~\triangleq~&\textit{HSR}(t_i)_{real}\xrightarrow{G_f}\textit{HSR}(t_{i+step})_{fake} \nonumber \\
    &\xrightarrow{G_b} \textit{HSR}(t_i)_{cycled},
    \label{eq:cycle1_definition}
\end{align}
where subscripts $fake$ and $cycled$ represent data generated from $real$ and data generated from $fake$, respectively; $G_f$ and $G_b$ indicate $\textit{Generator}_{forward}$ and $\textit{Generator}_{backward}$ in~\Cref{fig:schematic}, respectively.
$\textit{Cycle}~2$, which predicts the present from the known future and predicts the future again from the predicted present, is defined as follows:
\begin{align}
    \textit{Cycle}~2~\triangleq~&\textit{HSR}(t_{i+step})_{real}\xrightarrow{G_b}\textit{HSR}(t_{i})_{fake} \nonumber \\
    &\xrightarrow{G_f} \textit{HSR}(t_{i+step})_{cycled},
    \label{eq:cycle2_definition}
\end{align}
where variables are identical to variables in~\Cref{eq:cycle1_definition}.
In addition, $\textit{Discriminator}(t_{i+step})$ and $\textit{Discriminator}(t_i)$ are denoted as $D(t_{i+step})$ and $D(t_{i})$, respectively.
$\delta(t_{i+step})_{real}$, $\delta(t_{i+step})_{fake}$, $\delta(t_i)_{real}$, and $\delta(t_i)_{fake}$, the outputs of $D(t_{i+step})$ and $D(t_{i})$, are denoted as follows:
\begin{align}
    \delta(t_{i+step})_{real}&\triangleq D(t_{i+step})\big(\textit{HSR}(t_{i+step})_{real}\big), \label{eq:d(hsp(i+step))} \\
    \delta(t_{i+step})_{fake}&\triangleq D(t_{i+step})\big(\textit{HSR}(t_{i+step})_{fake}\big), \label{eq:d(g_f(hsp(i)))} \\
    \delta(t_i)_{real}&\triangleq D(t_{i})\big(\textit{HSR}(t_{i})_{real}\big), \label{eq:d(hsp(i))} \\
    \delta(t_i)_{fake}&\triangleq D(t_{i})\big(\textit{HSR}(t_{i})_{fake}\big). \label{eq:d(g_b(hsp(i+step)))}
\end{align}
And the $L1$ norms $L1(t_{i+step})_{fake}$, $L1(t_{i+step})_{cycled}$, $L1(t_{i})_{fake}$, and $L1(t_i)_{cycled}$ are denoted as follows:
\begin{align}
    L1&(t_{i+step})_{fake} \nonumber \\
    &~~~~~\triangleq\big\|\textit{HSR}(t_{i+step})_{fake}-\textit{HSR}(t_{i+step})_{real}\big\|_{1}, \label{eq:l1(i+step)_fake} \\
    L1&(t_{i+step})_{cycled} \nonumber \\
    &~~~~~\triangleq\big\|\textit{HSR}(t_{i+step})_{cycled}-\textit{HSR}(t_{i+step})_{real}\big\|_{1}, \label{eq:l1(i+step)_cycled} \\
    L1&(t_{i})_{fake} \nonumber \\
    &~~~~~\triangleq\big\|\textit{HSR}(t_{i})_{fake}-\textit{HSR}(t_{i})_{real}\big\|_{1}, \label{eq:l1(i)_fake} \\
    L1&(t_i)_{cycled} \nonumber \\
    &~~~~~\triangleq\big\|\textit{HSR}(t_{i})_{cycled}-\textit{HSR}(t_i)_{real}\big\|_{1}. \label{eq:l1(i)_cycled}
\end{align}
Note that Equations (\ref{eq:d(hsp(i+step))})--(\ref{eq:d(g_b(hsp(i+step)))}) are used in~\Cref{subsubsec:adv_loss}; Equations (\ref{eq:l1(i+step)_fake})--(\ref{eq:l1(i)_cycled}) are used in~\Cref{subsubsec:cycle_loss} and~\ref{subsubsec:connection_loss}.

\subsection{Optimization} 
\label{subsec:optimization}
PCT-CycleGAN receives $\textit{HSR}_i$ as input and distributes $\textit{HSR}(t_i)_{real}$ and $\textit{HSR}(t_{i+step})_{real}$ to $\textit{Cycle}~1$ and $\textit{Cycle}~2$, respectively.
With the complementary optimization of $\textit{Cycle}~1$ and $\textit{Cycle}~2$, the model obtains robust temporal causality.
Our proposed objective functions can be formulated into four kinds of terms,~\ie, adversarial loss, cycle-consistency loss, connection loss, and torrential loss.
The description of the four losses continues in~\Cref{subsubsec:adv_loss},~\ref{subsubsec:cycle_loss},~\ref{subsubsec:connection_loss}, and~\ref{subsubsec:torrential_loss} respectively.

\subsubsection{Adversarial Loss}
\label{subsubsec:adv_loss}
We apply adversarial losses to match the distributions of $\textit{HSR}(t_{i+step})_{real}$ and $\textit{HSR}(t_{i+step})_{fake}$. (\textit{Adversarial Loss (forward)} in~\Cref{fig:schematic}.)
Likewise, we also apply adversarial losses to match the distribution of $\textit{HSR}(t_{i})_{real}$ and $\textit{HSR}(t_{i})_{fake}$. (\textit{Adversarial Loss (backward)} in~\Cref{fig:schematic}.)
In particular, we adopt the adversarial losses of LSGANs and extend them for two complementary cycles~\cite{mao2017, mao2019}.
In other words, our adversarial losses are as follows:
\begin{align}
    \mathcal{L}_{adv}&\big(D(t_{i+step})\big) \nonumber \\
    =~&\frac{1}{2}\mathbb{E}_{\textit{HSR}(t_{i+step})_{real}\sim p(t_{i+step})}\Big[\big(\delta(t_{i+step})_{real}-1\big)^2\Big] \nonumber \\
    +~&\frac{1}{2}\mathbb{E}_{\textit{HSR}(t_{i})_{real}\sim p(t_i)}\Big[\big(\delta(t_{i+step})_{fake}\big)^2\Big], \label{eq:loss_adv_d(t_{i+step})} \\
    \mathcal{L}_{adv}&\big(G_f\big) \nonumber \\
    =~&\frac{1}{2}\mathbb{E}_{\textit{HSR}(t_{i})_{real}\sim p(t_i)}\Big[\big(\delta(t_{i+step})_{fake}-1\big)^2\Big], \label{eq:loss_adv_g_f} \\
    \mathcal{L}_{adv}&\big(D(t_{i})\big) \nonumber \\
    =~&\frac{1}{2}\mathbb{E}_{\textit{HSR}(t_{i})_{real}\sim p(t_{i})}\Big[\big(\delta(t_{i})_{real}-1\big)^2\Big] \nonumber \\
    +~&\frac{1}{2}\mathbb{E}_{\textit{HSR}(t_{i+step})_{real}\sim p(t_{i+step})}\Big[\big(\delta(t_{i})_{fake}\big)^2\Big], \label{eq:loss_adv_d(t_{i})} \\
    \mathcal{L}_{adv}&\big(G_b\big) \nonumber \\
    =~&\frac{1}{2}\mathbb{E}_{\textit{HSR}(t_{i+step})_{real}\sim p(t_{i+step})}\Big[\big(\delta(t_{i})_{fake}-1\big)^2\Big], \label{eq:loss_adv_g_b}
\end{align}
where $p(t_{i+step})$ and $p(t_i)$ indicate distributions of $\textit{HSR}(t_{i+step})_{real}$ and $\textit{HSR}(t_{i})_{real}$, respectively.

In our preliminary experiments, training using the adversarial losses of regular GANs was not successful.
Also, training using WGAN-GP required more time.
So the adversarial losses of LSGANs were finally chosen.

\subsubsection{Cycle-Consistency Loss}
\label{subsubsec:cycle_loss}
We apply the cycle-consistency loss into the time-series domain to create temporal causality~\cite{zhu2017}.
In other words, the cycle-consistency loss is used to satisfy the following conditions:
\begin{align}
    \textit{HSR}(t_{i+step})_{cycled}&\approx\textit{HSR}(t_{i+step})_{real}, \label{eq:condtion_t(i+step)_c_r} \\
    \textit{HSR}(t_{i})_{cycled}&\approx\textit{HSR}(t_i)_{real}. \label{eq:condtion_ti_c_r}
\end{align}
Therefore, the cycle-consistency loss, represented in purple in~\Cref{fig:schematic}, is as follows:
\begin{align}
    \mathcal{L}_{cyc}&\big(G_f, G_b\big) \nonumber \\
    =~&\mathbb{E}_{\textit{HSR}(t_{i+step})_{real}\sim p(t_{i+step})}\Big[L1(t_{i+step})_{cycled}\Big] \nonumber \\
    +~&\mathbb{E}_{\textit{HSR}(t_i)_{real}\sim p(t_i)}\Big[L1(t_i)_{cycled}\Big]. \label{eq:loss_cycle}
\end{align}
Here, $p(t_{i+step})$ and $p(t_i)$ indicate distributions of $\textit{HSR}(t_{i+step})_{real}$ and $\textit{HSR}(t_{i})_{real}$, respectively.

\subsubsection{Connection Loss}
\label{subsubsec:connection_loss}
We apply novel connection losses with adversarial losses to satisfy the following conditions:
\begin{align}
    \textit{HSR}(t_{i+step})_{real}&\approx\textit{HSR}(t_{i+step})_{fake}, \label{eq:condition_t(i+step)_r_f} \\
    \textit{HSR}(t_{i})_{real}&\approx\textit{HSR}(t_i)_{fake}. \label{eq:condition_ti_r_f}
\end{align}
To satisfy Equations~(\ref{eq:condition_t(i+step)_r_f}) and~(\ref{eq:condition_ti_r_f}), the connection loss (forward) $\mathcal{L}_{con}(G_f)$ and connection loss (backward) $\mathcal{L}_{con}(G_b)$ are as follows:
\begin{align}
    \mathcal{L}_{con}&\big(G_f\big) \nonumber \\
    =~&\mathbb{E}_{\textit{HSR}_{i}\sim(p(t_i),~p(t_{i+step}))}\Big[L1(t_{i+step})_{fake}\Big], \label{eq:loss_connection_g_f} \\
    \mathcal{L}_{con}&\big(G_b\big) \nonumber \\
    =~&\mathbb{E}_{\textit{HSR}_{i}\sim(p(t_i),~p(t_{i+step}))}\Big[L1(t_{i})_{fake}\Big], \label{eq:loss_connection_g_b}
\end{align}
where $p(t_{i})$ and $p(t_{i+step})$ indicate distributions of $\textit{HSR}(t_{i})_{real}$ and $\textit{HSR}(t_{i+step})_{real}$, respectively.
Also, in~\Cref{fig:schematic}, $\mathcal{L}_{con}(G_f)$ and $\mathcal{L}_{con}(G_b)$ are represented in blue and red, respectively.

In our preliminary experiments, using only adversarial losses and cycle-consistency loss did not capture the changes over time well.
However, the identity loss, additionally adopted in the original CycleGAN to improve performance~\cite{zhu2017}, could not be used in our experiments because it breaks temporal causality.
For example, $G_f$ should always be a mapping from the present to the future, but if the identity loss is added, it can also represent a mapping from the present to the present.
Instead of the identity loss, we added connection losses, inspired by Pix2Pix~\cite{isola2017}, to help $G_f$ and $G_b$ capture the temporal changes.

\begin{figure*}[t]
\begin{lstlisting}[basicstyle=\tiny]
import tensorflow as tf

class Torrential(tf.keras.losses.Loss):
    def __init__(self, threshold, epsilon = 0):
        super(Torrential, self).__init__()
        self.threshold = threshold
        self.epsilon = epsilon

    def call(self, y_true, y_pred):
        binary_boolean_true = tf.where(condition=(y_true >= self.threshold), x=1, y=0)
        binary_boolean_pred = tf.where(condition=(y_pred >= self.threshold), x=1, y=0)
        num_hit_mask = tf.reduce_sum(input_tensor=(binary_boolean_true & binary_boolean_pred))
        num_total_mask = tf.reduce_sum(input_tensor=(binary_boolean_true | binary_boolean_pred))
        if num_total_mask == 0:
            return tf.cast(x=0, dtype=tf.float32) + self.epsilon
        else:
            csi = num_hit_mask / num_total_mask
            return tf.cast(x=tf.math.abs(x= (1 - csi)), dtype=tf.float32) + self.epsilon
\end{lstlisting}
\vspace{-5mm}
\caption{Pseudo-code of torrential loss in the TensorFlow 2.x style.}
\vspace{-4.5mm}
\label{fig:pseudo_torrential}
\end{figure*}
\subsubsection{Torrential Loss}
\label{subsubsec:torrential_loss}
We apply torrential losses to cover exceptional heavy rain events well.
The torrential loss aims to maximize the CSI of generated HSR~\cite{schaefer1990}.
According to design criteria, the threshold of CSI is constrained to a high value above 25mm/h.
Therefore, the torrential loss (forward) $\mathcal{L}_{tor}(G_f)$ and torrential loss (backward) $\mathcal{L}_{tor}(G_b)$ are as follows:
\begin{align}
    \mathcal{L}_{tor}&\big(G_f\big) \nonumber \\
    =~&\left\{
        \begin{array}{ll}
            0,~\text{if $\textit{CSI}_{\theta}(t_{i+step})_{fake}$ does not exist}, \\
            \mathbb{E}_{\textit{HSR}_{i}\sim(p(t_i),~p(t_{i+step}))}\Big[1 - \textit{CSI}_{\theta}(t_{i+step})_{fake}\Big],~\text{otherwise},
        \end{array}
    \right. \label{eq:loss_torrential_g_f} \\
    \mathcal{L}_{tor}&\big(G_b\big) \nonumber \\
    =~&\left\{
        \begin{array}{ll}
            0,~\text{if $\textit{CSI}_{\theta}(t_{i})_{fake}$ does not exist}, \\
            \mathbb{E}_{\textit{HSR}_{i}\sim(p(t_i),~p(t_{i+step}))}\Big[1 - \textit{CSI}_{\theta}(t_{i})_{fake}\Big],~\text{otherwise}.
        \end{array}
    \right. \label{eq:loss_torrential_g_b}
\end{align}
Here, $\theta$, $\textit{CSI}_{\theta}(t_{i+step})_{fake}$, and $\textit{CSI}_{\theta}(t_{i})_{fake}$ are the threshold, CSI of $\textit{HSR}(t_{i+step})_{fake}$, and CSI of $\textit{HSR}(t_i)_{fake}$, respectively.
$p(t_{i})$ and $p(t_{i+step})$ indicate distributions of $\textit{HSR}(t_{i})_{real}$ and $\textit{HSR}(t_{i+step})_{real}$, respectively.

The concept of CSI~\cite{schaefer1990}, a de facto standard metric for weather forecasting, may be unfamiliar to non-meteorologists.
Thus, we provide the pseudo-code of torrential loss to help readers understand in~\Cref{fig:pseudo_torrential}.

\subsubsection{Total Objective Function}
\label{subsubsec:objective_function}
According to Equations (\ref{eq:loss_adv_d(t_{i+step})})--(\ref{eq:loss_adv_g_b}), (\ref{eq:loss_cycle}), and (\ref{eq:loss_connection_g_f})--(\ref{eq:loss_torrential_g_b}), total objective functions are as follows:
\begin{align}
    \min_{D(t_{i+step})}:~\mathcal{L}\big(D&(t_{i+step})\big) \nonumber \\
    =~&\mathcal{L}_{adv}\big(D(t_{i+step})\big), \label{eq:objective_d_i_step} \\
    \min_{G_f}\hspace{3mm}:~\mathcal{L}\big(G&_f\big) \nonumber \\
    =~&\mathcal{L}_{adv}\big(G_f\big)+\lambda_{cyc}\mathcal{L}_{cyc}\big(G_f, G_b\big) \nonumber \\
    +~&\lambda_{con}\mathcal{L}_{con}\big(G_f\big)+\lambda_{tor}\mathcal{L}_{tor}\big(G_f\big), \label{eq:objective_g_f} \\
    \min_{D(t_{i})}\hspace{2mm}:~\mathcal{L}\big(D&(t_{i})\big) \nonumber \\
    =~&\mathcal{L}_{adv}\big(D(t_{i})\big), \label{eq:objective_d_i} \\
    \min_{G_b}\hspace{2.5mm}:~\mathcal{L}\big(G&_b\big) \nonumber \\
    =~&\mathcal{L}_{adv}\big(G_b\big)+\lambda_{cyc}\mathcal{L}_{cyc}\big(G_f, G_b\big) \nonumber \\
    +~&\lambda_{con}\mathcal{L}_{con}\big(G_b\big)+\lambda_{tor}\mathcal{L}_{tor}\big(G_b\big), \label{eq:objective_g_b}
\end{align}
where $\lambda_{cyc}\in\mathbb{R}^{+}$, $\lambda_{con}\in\mathbb{R}^{+}$, and $\lambda_{tor}\in\mathbb{R}^{+}$ control their relative importance.
Our objective functions satisfy Equations (\ref{eq:condtion_t(i+step)_c_r}), (\ref{eq:condtion_ti_c_r}), (\ref{eq:condition_t(i+step)_r_f}), and (\ref{eq:condition_ti_r_f}); thus, our model achieves the following final goals:
\begin{align}
    \textit{HSR}(t_{i+step})_{real}&\approx\textit{HSR}(t_{i+step})_{fake} \nonumber \\
    &\approx\textit{HSR}(t_{i+step})_{cycled}, \label{eq:final_goal_i_step}\\
    \textit{HSR}(t_{i})_{real}&\approx\textit{HSR}(t_{i})_{fake} \nonumber \\
    &\approx\textit{HSR}(t_{i})_{cycled}. \label{eq:final_goal_i}
\end{align}
These goals guarantee the temporal causality of $\textit{Cycle}~1$ and $\textit{Cycle}~2$.

\begin{figure}[t]
    \centering
    \includegraphics[width=0.51\columnwidth]{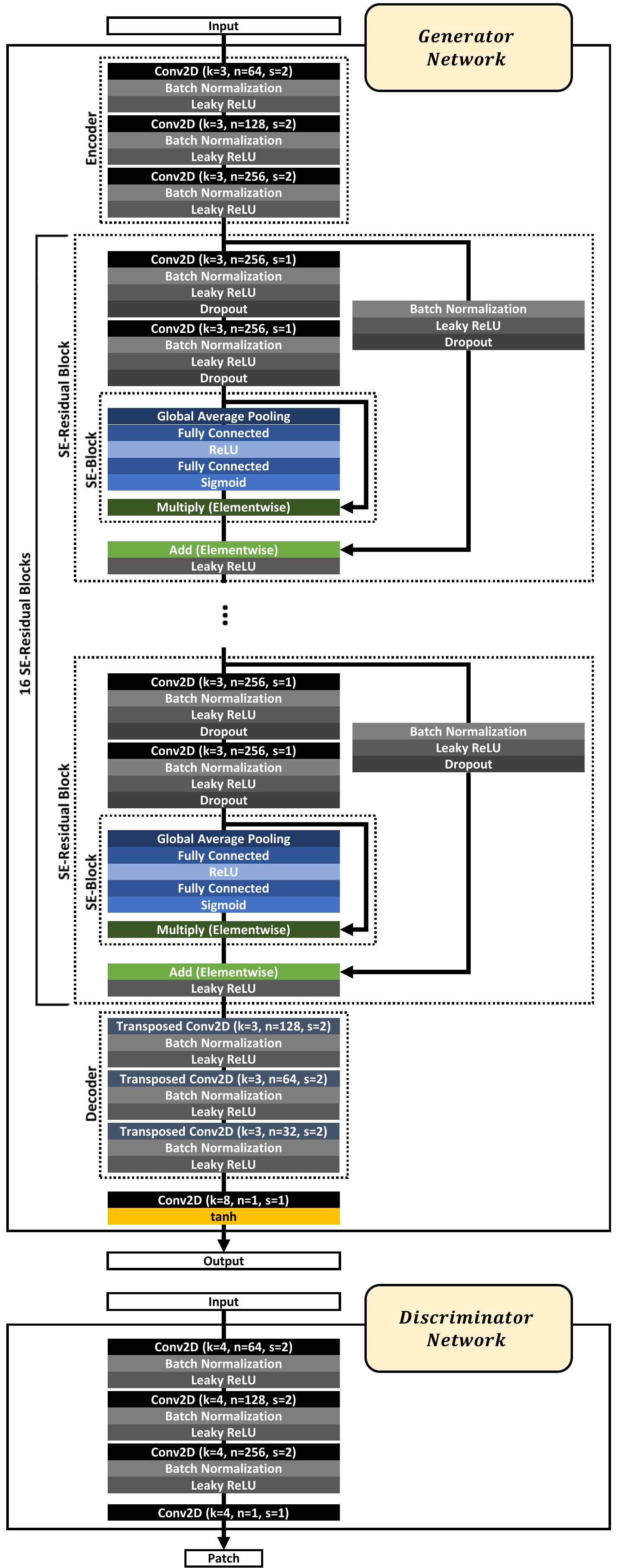}
    \vspace{-3mm}
    \caption{
    Network architectures of generators and discriminators.
    The generators consist of the encoder, squeeze-and-excitation (SE)-residual blocks~\cite{he2016, hu2018}, and decoder.
    The discriminators use the architecture of PatchGANs~\cite{li2016}.
    Here, $k$, $n$, and $s$ in the convolution layers and the transposed convolution layers are the kernel size, the number of feature maps, and the stride, respectively.
    }
    \vspace{-5mm}
    \label{fig:network_architecture}
\end{figure}
\subsection{Network Architecture}
\label{subsec:architecture}
Our generators $G_f$ and $G_b$ have the same network architectures.
And our discriminators $D(t_{i+step})$ and $D(t_i)$ have the same network architectures.
The network architectures of our discriminators and generators are shown in~\Cref{fig:network_architecture}.

\subsubsection{Generator Architecture}
\label{subsubsec:generator_architecture}
The generator network has an encoder, $16$ squeeze-and-excitation (SE)-residual blocks~\cite{he2016, hu2018}, and a decoder.
The encoder reduces the size of the input HSR and extracts $256$ features.
The SE-residual block is a core part of the generator network, a combination of the residual block and the SE-block.
We add the dropout layers in the residual blocks for robustness.
We use a small number of residual blocks, so we do not use the bottleneck block according to the design policy of the original ResNet.
The decoder is paired with the encoder and restores the size of the reduced HSR data.
The convolution layer just before the output is for fine resizing, and $\tanh$ is used for the final activation function.

In our preliminary experiments, U-Net architecture~\cite{ronneberger2015}, which is widely used in the meteorological field, failed to learn as the data distributions became more diverse.
We estimated that the failure of U-Net is due to the skip-connection techniques using the concatenation layers that copy and paste channels.
Instead, we used the residual blocks which use elementwise sum operations for skip-connection techniques~\cite{he2016}.
Also, the addition of the SE-blocks increased the quality of the generated data~\cite{hu2018}.

\subsubsection{Discriminator Architecture}
\label{subsubsec:discriminator_architecture}
The discriminator network uses the architecture of PatchGANs~\cite{li2016}.
In other words, our discriminator network classifies whether the output data patch is real or fake.
Such the patch-unit discriminator is lighter than the full-data discriminator~\cite{isola2017}.
We use the $31\times31$ patch size to distinguish.
However, the patch size can be flexibly changed according to the size of the input data.

\section{Results}
\label{sec:results}
We first describe the datasets used and training details.
We then demonstrate the superiority of PCT-CycleGAN through qualitative and quantitative evaluations.
Finally, we discuss the limitations of the proposed model.

\begin{table}[t]
\centering
\caption{Three different HSR datasets information.}
\label{tab:dataset}
\vspace{-3mm}
\resizebox{\columnwidth}{!}{%
\begin{tabular}{c|ccc}
\hline
Dataset      & Seoul       & Jeju Island & Daegwallyeong \\ \hline \hline
K\"{o}ppen-Geiger          & \multirow{2}{*}{Dwa} & \multirow{2}{*}{Cfa} & \multirow{2}{*}{Dfb}   \\
Climate Classification &                      &                      &                        \\ \hline
Spatial Area           & 240km$\times$240km        & 240km$\times$240km        & 240km$\times$240km          \\ \hline
Spatial Resolution     & 1km                  & 1km                  & 1km                    \\ \hline
Temporal Resolution    & 5 minutes            & 5 minutes            & 5 minutes              \\ \hline
\multirow{2}{*}{Train Dataset} & 2021-07-01 $\sim$ & 2021-07-01 $\sim$ & 2021-07-01 $\sim$ \\
                       & 2021-12-31           & 2021-12-31           & 2021-12-31             \\ \hline
\multirow{2}{*}{Test Dataset}  & 2022-07-01 $\sim$ & 2022-08-01 $\sim$ & 2022-08-01 $\sim$ \\
                       & 2022-09-30           & 2022-09-30           & 2022-09-30 \\ \hline
\end{tabular}%
}
\vspace{-4mm}
\end{table}
\subsection{Datasets}
\label{subsec:dataset}
As shown in~\Cref{tab:dataset}, we use HSR datasets from three regions with different K\"{o}ppen-Geiger climate classifications~\cite{beck2018}.
The default of $step$, the interval between paired data, is $2$.
In other words, the default time interval between paired data is $10$ minutes.

\subsection{Training Details}
\label{subsec:training_details}
In the data preprocessing process, we remove the pair $\textit{HSR}_i$, where $\textit{HSR}(t_i)_{real}==\textit{HSR}(t_{i+step})_{real}$.
This is to focus on the extinction and development of radar echoes.
Then, the refined dataset is normalized to $[-1,~1]$.
In the generator and discriminator networks, we set the momentum for the moving average of batch normalization to $0.8$ and the negative slope of Leaky ReLU to $0.2$.
We also set the drop rate of the dropout layer to $0.4$.
As the optimizer, we use the Adam with momentum parameters $\beta_{1}=0.5$ and $\beta_{2}=0.999$~\cite{kingma2014}.
The learning rate and batch size are set to $0.0002$ and $16$, respectively.
We set $\theta=30$ in torrential losses (\ref{eq:loss_torrential_g_f}) and (\ref{eq:loss_torrential_g_b}).
In our objective functions (\ref{eq:objective_g_f}) and (\ref{eq:objective_g_b}), we set $\lambda_{cyc}=10$, $\lambda_{con}=10$, and $\lambda_{tor}=100$.
We use the NVIDIA A100 Tensor Core GPU for training.

\subsection{Evaluation}
\label{subsec:evaluation}
In this section, comparative evaluations and ablation studies are conducted.
In our experiments, PCT-CycleGAN performs iterative forecasting in $10$-minute increments.

\subsubsection{Comparison against Baselines}
\label{subsubsec:comparison}
Our method is compared both qualitatively and quantitatively to the three baselines.
The first baseline is MAPLE~\cite{cho2021, turner2004}, one of the best QPF models.
Note that MAPLE we used is a $2021$ version that has gone through many years of optimization and variational computation method improvement~\cite{cho2021}.
The second baseline is ConvLSTM based on RNNs~\cite{shi2015}.
ConvLSTM is the most commonly used learning method for precipitation nowcasting.
The third baseline is MetNet-2~\cite{espeholt2022}, the latest sophisticated precipitation nowcasting method.
For qualitative evaluation, we use the CSI as a metric~\cite{schaefer1990}.
For quantitative evaluation, we use PSNR and SSIM as metrics~\cite{wang2004}.
CSI, PSNR, and SSIM are all the higher the better.

\begin{table}[t]
\centering
\caption{Quantitative comparisons of precipitation nowcasting in Seoul.}
\label{tab:seoul_case}
\vspace{-3mm}
\resizebox{\columnwidth}{!}{%
\begin{tabular}{c|c|cccccc}
\hline
Method                    & Metric & $+20min$ & $+40min$ & $+60min$ & $+80min$ & $+100min$ & $+120min$ \\ \hline \hline
\multirow{2}{*}{PCT-CycleGAN}     & PSNR   & \textbf{25.558}   & \textbf{24.641}   & 24.462   & 24.192   & 24.125    & \textbf{24.104}    \\
                          & SSIM   & \textbf{0.898}    & \textbf{0.885}    & 0.870    & 0.854    & 0.844     & 0.842     \\ \hline
\multirow{2}{*}{MAPLE}    & PSNR   & 25.050   & 23.517   & 23.351   & 23.057   & 23.981    & 23.818    \\
                          & SSIM   & 0.897    & 0.864    & 0.847    & 0.837    & 0.830     & 0.836     \\ \hline
\multirow{2}{*}{ConvLSTM} & PSNR   & 22.599   & 20.825   & 20.098   & 20.542   & 20.223    & 19.593    \\
                          & SSIM   & 0.858    & 0.850    & 0.843    & 0.841    & 0.829     & 0.821     \\ \hline
\multirow{2}{*}{MetNet-2} & PSNR   & 24.078   & 23.453   & \textbf{25.744}   & \textbf{25.125}   & \textbf{25.990}    & 23.551    \\
                          & SSIM   & 0.895    & 0.884    & \textbf{0.898}    & \textbf{0.900}    & \textbf{0.897}     & \textbf{0.893}      \\ \hline
\end{tabular}%
}
\vspace{-3mm}
\end{table}
\begin{table}[t]
\centering
\caption{Quantitative comparisons of precipitation nowcasting in Jeju Island.}
\label{tab:jeju_case}
\vspace{-3mm}
\resizebox{\columnwidth}{!}{%
\begin{tabular}{c|c|cccccc}
\hline
Method                    & Metric & $+20min$ & $+40min$ & $+60min$ & $+80min$ & $+100min$ & $+120min$ \\ \hline \hline
\multirow{2}{*}{PCT-CycleGAN}& PSNR  & 31.389 & \textbf{32.121} & \textbf{32.268} & 28.776 & \textbf{28.748} & \textbf{27.768} \\
                          & SSIM   & 0.956 & \textbf{0.949} & \textbf{0.930} & 0.903 & \textbf{0.899} & \textbf{0.900} \\ \hline
\multirow{2}{*}{MAPLE}    & PSNR   & 30.528 & 27.431 & 26.266 & 25.467 & 24.922 & 24.944 \\
                          & SSIM   & 0.961 & 0.942 & 0.905 & 0.876 & 0.864 & 0.866 \\ \hline
\multirow{2}{*}{ConvLSTM} & PSNR   & \textbf{33.244} & 31.697 & 31.442 & \textbf{29.935} & 27.775 & 26.747 \\
                          & SSIM   & \textbf{0.965} & 0.943 & 0.922 & \textbf{0.912} & 0.886 & 0.884 \\ \hline
\multirow{2}{*}{MetNet-2} & PSNR   & 27.505 &  28.532 &  29.013  & 28.976  & 28.678 & 27.746    \\
                          & SSIM   & 0.922   & 0.910 & 0.885 & 0.875  & 0.871   &  0.880     \\ \hline
\end{tabular}%
}
\vspace{-3mm}
\end{table}
\begin{table}[t]
\centering
\caption{Quantitative comparisons of precipitation nowcasting in Daegwallyeong.}
\label{tab:dae_case}
\vspace{-3mm}
\resizebox{\columnwidth}{!}{%
\begin{tabular}{c|c|cccccc}
\hline
Method                    & Metric & $+20min$ & $+40min$ & $+60min$ & $+80min$ & $+100min$ & $+120min$ \\ \hline \hline
\multirow{2}{*}{PCT-CycleGAN}& PSNR  & 28.839 & 28.728 & \textbf{28.711} & \textbf{28.968} & \textbf{29.876} & \textbf{27.799} \\
                          & SSIM   & 0.878 & 0.872 & 0.877 & 0.877 & \textbf{0.875} & \textbf{0.857} \\ \hline
\multirow{2}{*}{MAPLE}    & PSNR   & 28.507 & 27.991 & 27.865 & 27.920 & 27.937 & 27.349 \\
                          & SSIM   & 0.906 & 0.896 & \textbf{0.888} & 0.882 & 0.869 & 0.848 \\ \hline
\multirow{2}{*}{ConvLSTM} & PSNR   & 28.549 & 27.589 & 27.062 & 27.082 & 23.354 & 24.339 \\
                          & SSIM   & 0.904 & \textbf{0.908} & 0.812 & 0.802 & 0.851 & 0.849 \\ \hline
\multirow{2}{*}{MetNet-2} & PSNR   & \textbf{29.285} & \textbf{28.742} & 28.272 & 28.776 & 28.892 & 23.928 \\
                          & SSIM   & \textbf{0.915} & 0.881 & 0.884 & \textbf{0.906} & 0.870 & 0.812 \\ \hline
\end{tabular}%
}
\vspace{-4mm}
\end{table}
\begin{figure*}[t]
    \centering
    \includegraphics[width=0.88\textwidth]{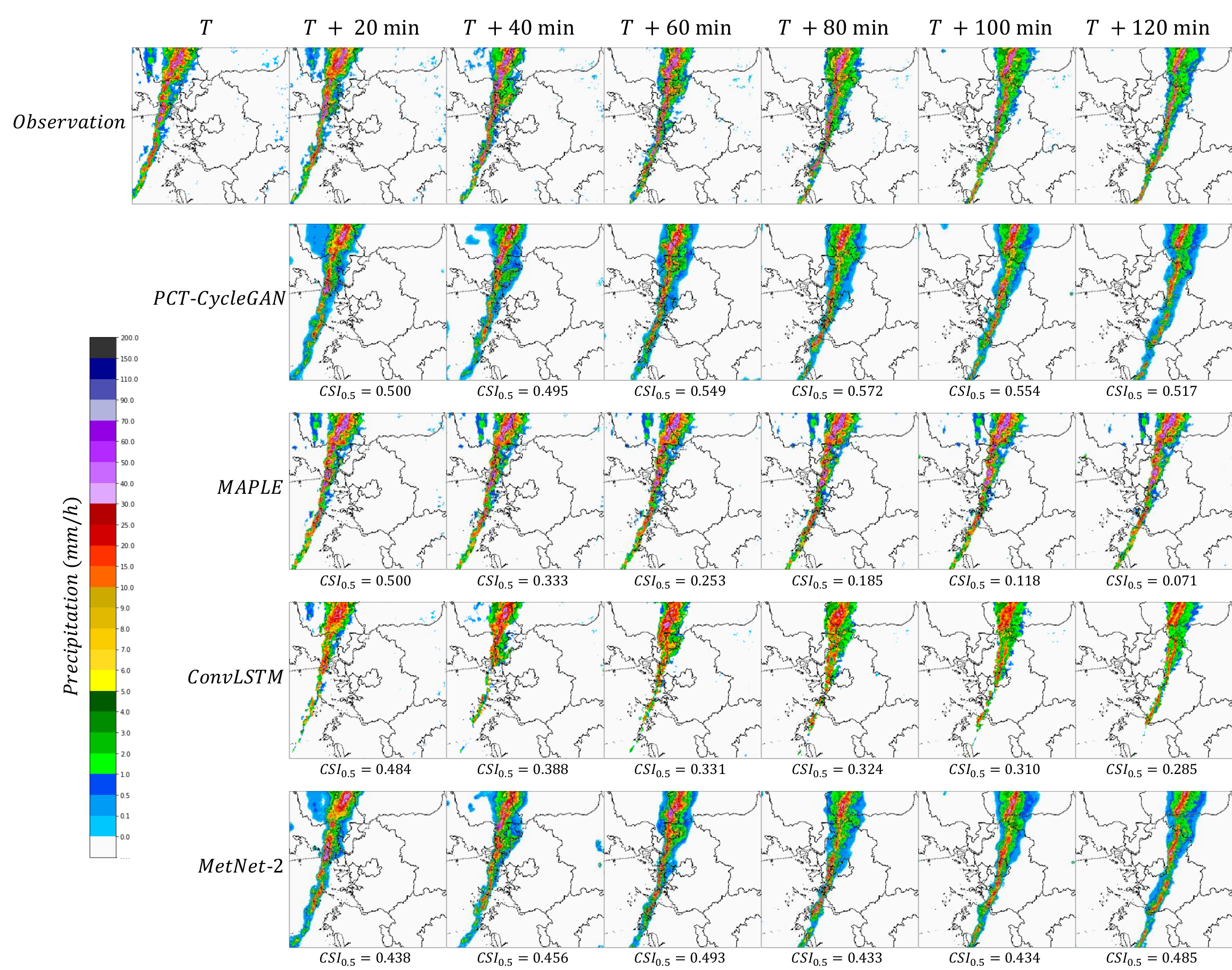}
    \vspace{-3mm}
    \caption{
    Qualitative comparisons of precipitation nowcasting in Seoul.
    The starting point \textit{T} is 2022-09-23 01:00 (UTC). 
    The threshold of the critical success index (CSI) is 0.5mm/h.
    }
    \vspace{-3mm}
    \label{fig:results_seoul}
\end{figure*}
\begin{figure*}[t]
    \centering
    \includegraphics[width=0.88\textwidth]{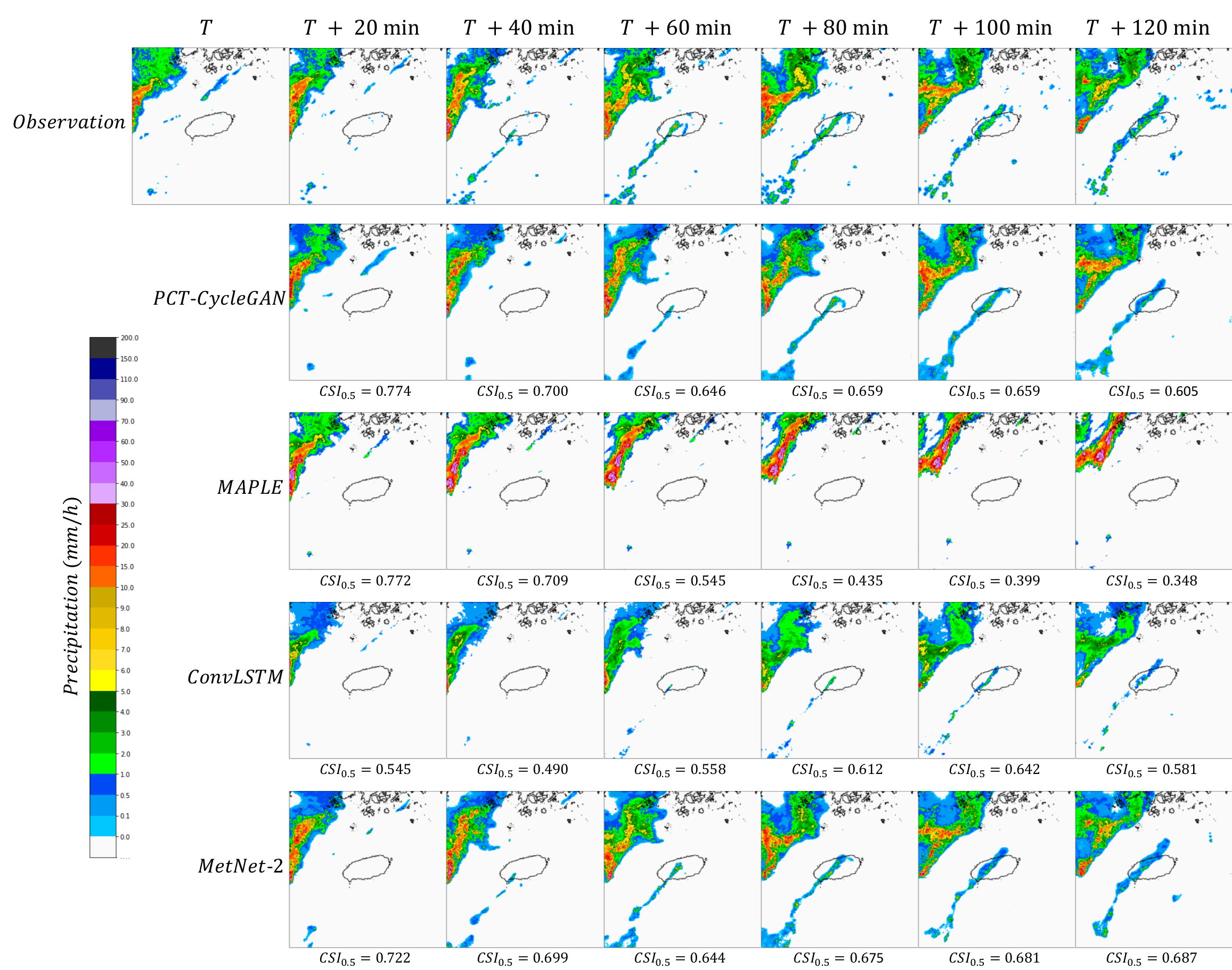}
    \vspace{-3mm}
    \caption{
    Qualitative comparisons of precipitation nowcasting in Jeju Island.
    The starting point \textit{T} is 2022-09-11 17:30 (UTC). 
    The threshold of the CSI is 0.5mm/h.
    }
    \vspace{-3mm}
    \label{fig:results_jeju}
\end{figure*}
\begin{figure*}[t]
    \centering
    \includegraphics[width=0.88\textwidth]{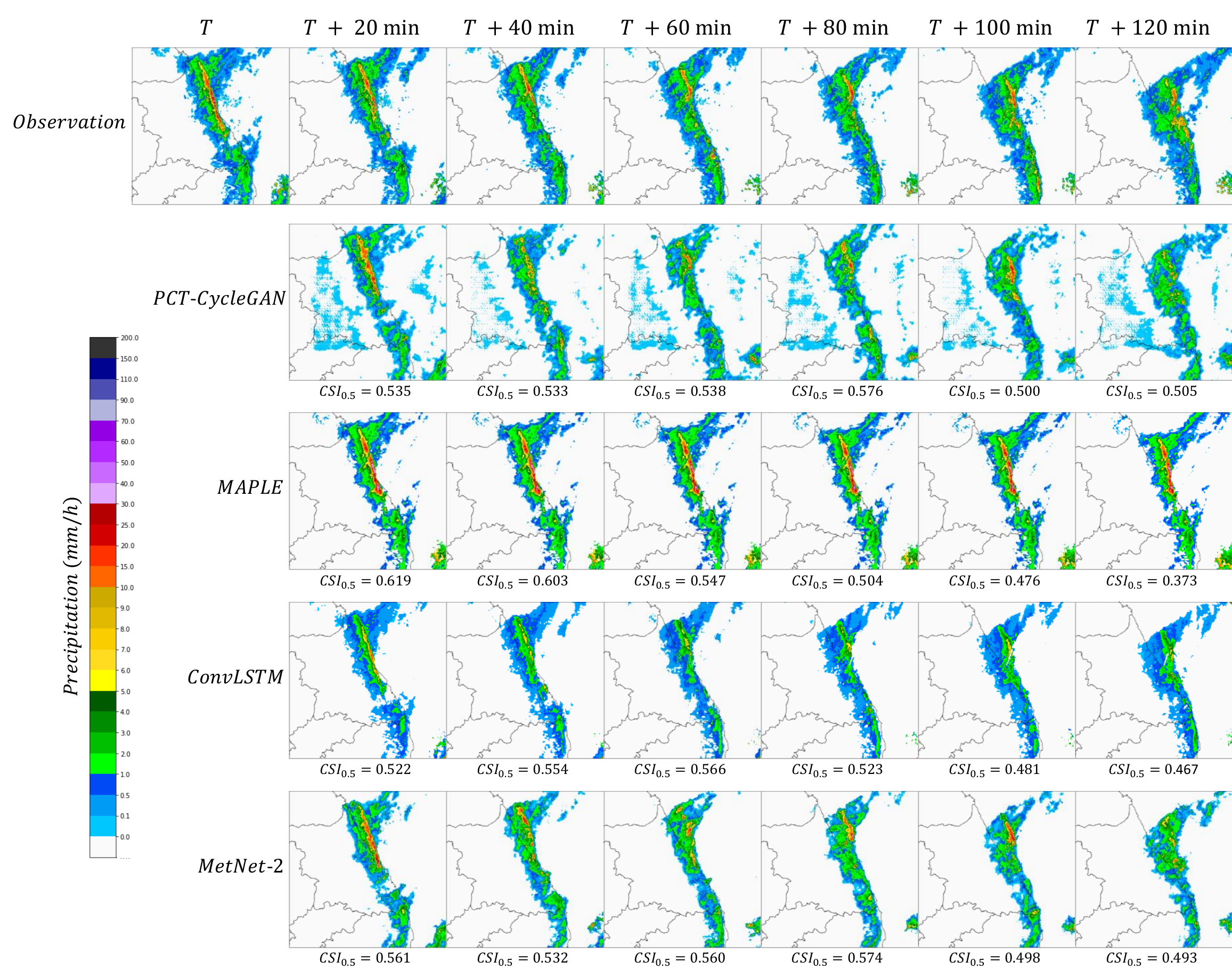}
    \vspace{-3mm}
    \caption{
    Qualitative comparisons of precipitation nowcasting in Daegwallyeong.
    The starting point \textit{T} is 2022-09-19 01:00 (UTC). 
    The threshold of the CSI is 0.5mm/h.
    }
    \vspace{-3mm}
    \label{fig:results_dae}
\end{figure*}
\vspace{0.5mm}
\noindent
$\bullet$~{\bf Seoul Dataset (Dwa). }
As shown in~\Cref{fig:results_seoul}, in the experimental results of the CSI-based qualitative evaluation, PCT-CycleGAN guaranteed the highest score in all time points.
The CSI of PCT-CycleGAN and MetNet-2 remained relatively uniform compared to the CSI of other baselines during a lead time of two hours.
In particular, PCT-CycleGAN predicted the extinction of radar echoes better than MAPLE.
And PCT-CycleGAN predicted the shape of the squall line better than ConvLSTM~\cite{zipser1977}.
As shown in~\Cref{tab:seoul_case}, PCT-CycleGAN scored the best in both PSNR and SSIM during a lead time of $40$ minutes.
After $40$ minutes, PCT-CycleGAN was slightly behind MetNet-2.

\vspace{0.5mm}
\noindent
$\bullet$~{\bf Jeju Island Dataset (Cfa). }
As shown in~\Cref{fig:results_jeju}, PCT-CycleGAN and MetNet-2 maintained great CSI for a lead time of $2$ hours.
As shown in~\Cref{tab:jeju_case}, PCT-CycleGAN scored the best in both PSNR and SSIM except for only $20$ and $80$ minutes. 

\vspace{0.5mm}
\noindent
$\bullet$~{\bf Daegwallyeong Dataset (Dfb). }
As shown in~\Cref{fig:results_dae}, MAPLE guaranteed the highest CSI during a lead time of $40$ minutes~\cite{cho2021}.
However, MAPLE could not express the decay of the strong precipitation echoes, and the performance continued to decrease over time.
From $80$ minutes, PCT-CycleGAN outperformed the others.
Some falsely generated precipitation echoes appeared in the results of PCT-CycleGAN, but they were under the CSI threshold.
MetNet-2 showed a poor prediction of weak precipitation echoes compared to PCT-CycleGAN but showed similar overall performance.
As shown~\Cref{tab:dae_case}, PCT-CycleGAN recorded the highest PSNR from $60$ minutes and the highest SSIM from $100$ minutes.

Due to various factors such as topography and wind, the performance of nowcasting may vary by case.
Nevertheless, it is encouraging that PCT-CycleGAN performs outstandingly in all datasets.

\begin{table}[t]
\centering
\caption{
Quantitative comparisons with and without connection loss.
}
\vspace{-3mm}
{\footnotesize
\begin{tabular}{c|cc}
\hline
Metric & With Connection Loss & Without Connection Loss \\ 
\hline \hline
PSNR & 22.881 & 20.829 \\
SSIM & 0.895 & 0.867  \\
\hline
\end{tabular}%
}
\label{tab:ablation_connection}
\vspace{-3mm}
\end{table}
\begin{table}[t]
\centering
\caption{
Quantitative comparisons with and without torrential loss.
}
\vspace{-3mm}
{\footnotesize
\begin{tabular}{c|cc}
\hline
Metric & With Torrential Loss & Without Torrential Loss \\ 
\hline \hline
PSNR & 27.577 & 23.599 \\
SSIM & 0.881 & 0.827  \\
\hline
\end{tabular}%
}
\label{tab:ablation_torrential}
\vspace{-4mm}
\end{table}
\begin{figure}[t]
    \centering
    \includegraphics[width=0.75\columnwidth]{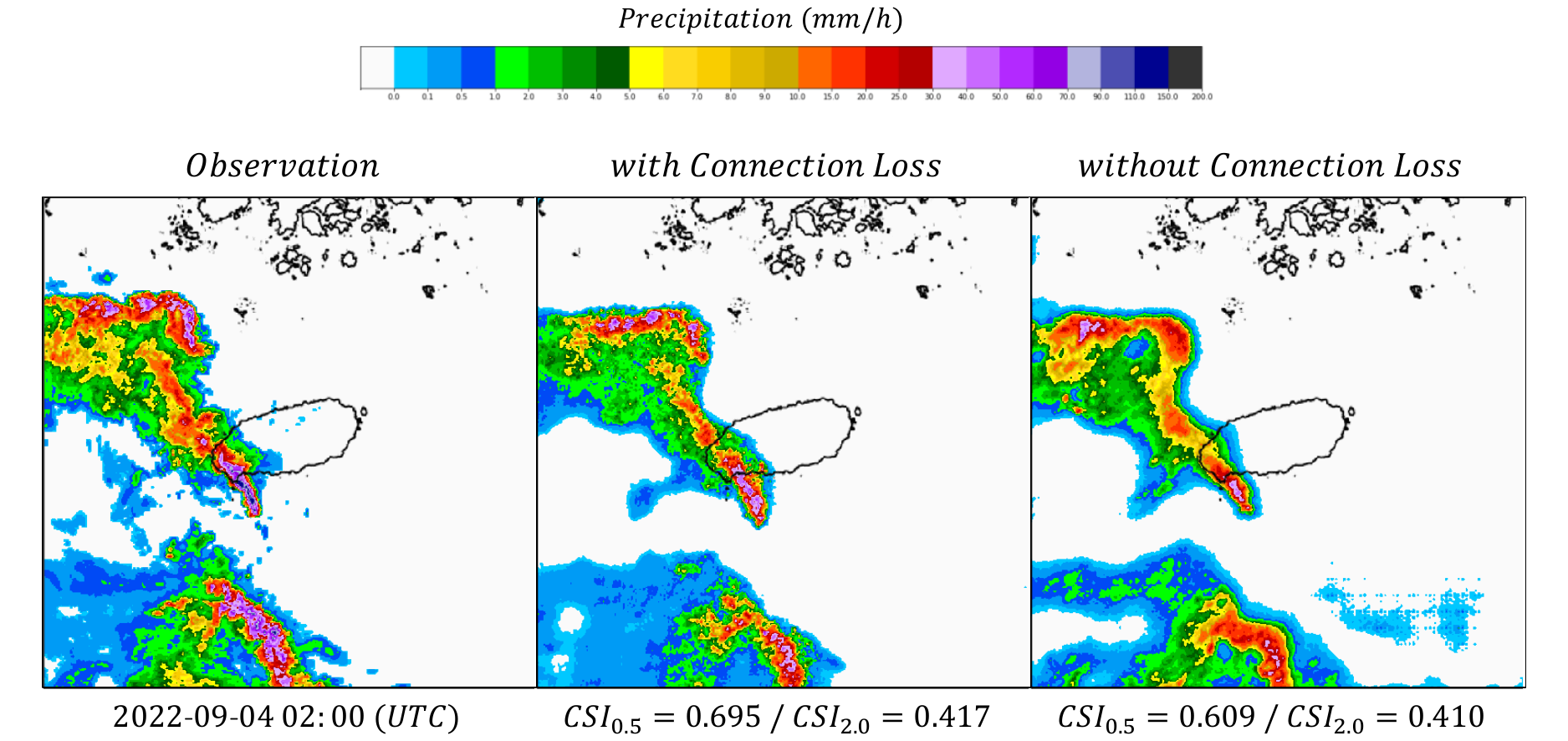}
    \vspace{-3mm}
    \caption{
    Qualitative comparisons with and without connection loss.
    Subscripts indicate threshold values (mm/h).
    }
    \label{fig:ablation_connection}
    \vspace{-3mm}
\end{figure}
\begin{figure}[t]
    \centering
    \includegraphics[width=0.75\columnwidth]{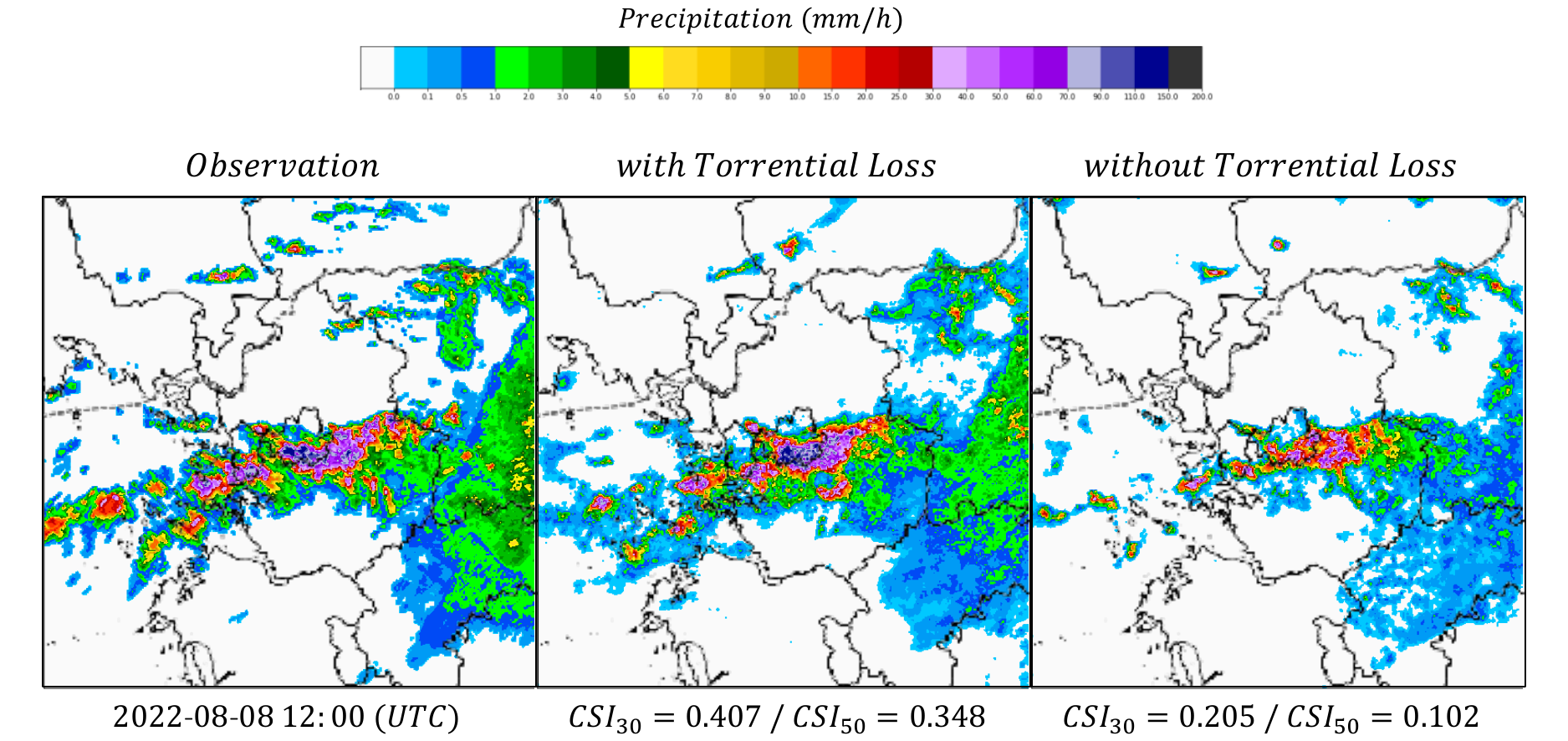}
    \vspace{-3mm}
    \caption{
    Qualitative comparisons with and without torrential loss.
    Subscripts indicate threshold values (mm/h).
    }
    \label{fig:ablation_torrential}
    \vspace{-3mm}
\end{figure}
\subsubsection{Ablation Study}
\label{subsubsec:ablation}
We conduct ablation studies with two clear goals.
The first goal is to prove the effectiveness of the proposed connection loss.
We remove only connection loss, with no changes in other experimental conditions.
As shown in~\Cref{fig:ablation_connection} and~\Cref{tab:ablation_connection}, connection loss makes temporal causality more robust, resulting in sharper outputs.
This is evidenced by the higher values of CSI, PSNR, and SSIM when using connection loss.

The second goal is to prove the effectiveness of the proposed torrential loss.
We remove only torrential loss, with no changes in other experimental conditions.
\Cref{fig:ablation_torrential} shows an uncommon heavy rain event in Seoul on August $8$, $2022$.
PCT-CycleGAN with torrential loss shows better performance compared to PCT-CycleGAN without torrential loss.
In particular, the difference in nowcasting performance is evident in areas with heavy rainfall.
The difference among the CSI is overwhelming.
As shown in~\Cref{tab:ablation_torrential}, it can be observed that the values of PSNR and SSIM are also higher when using torrential loss.

\begin{figure}[t]
    \centering
    \includegraphics[width=0.75\columnwidth]{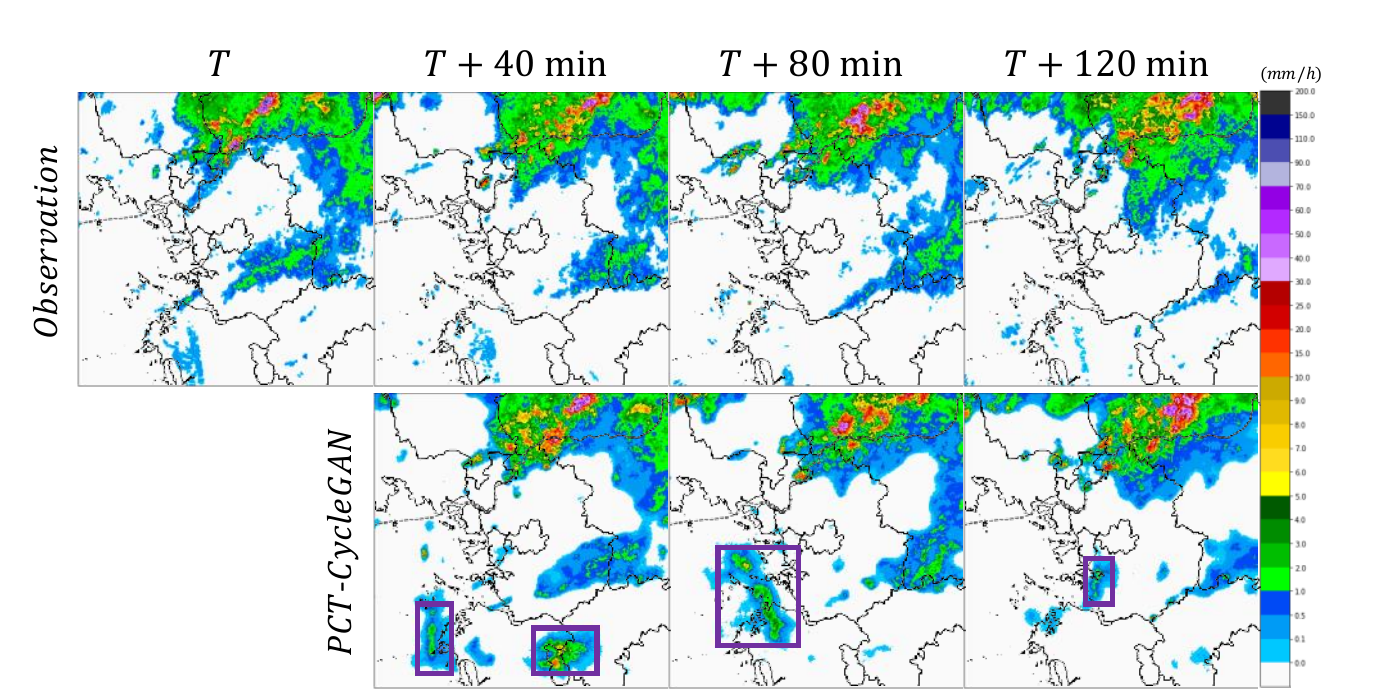}
    \vspace{-4mm}
    \caption{
    Failure case in Seoul.
    The starting point \textit{T} is 2022-08-07 15:00 (UTC).
    }
    \label{fig:limit_seoul}
    \vspace{-3mm}
\end{figure}
\begin{figure}[t]
    \centering
    \includegraphics[width=0.75\columnwidth]{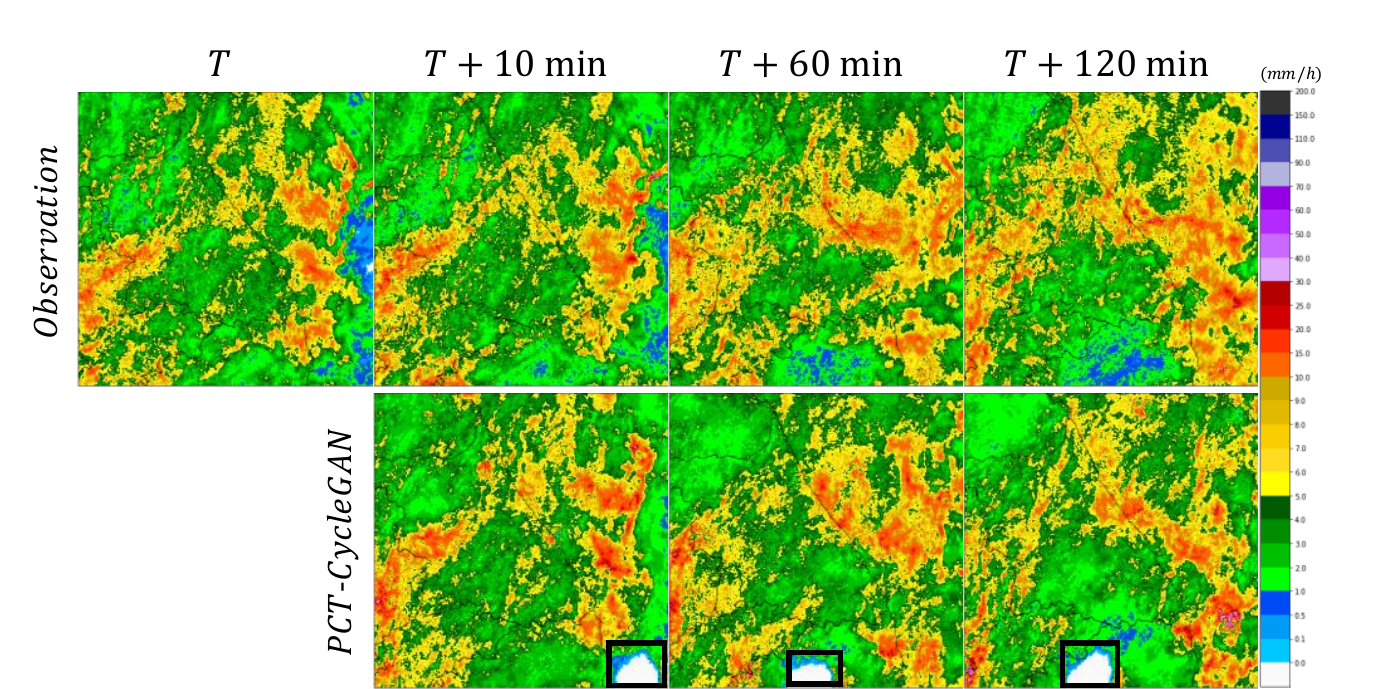}
    \vspace{-4mm}
    \caption{
    Failure case in Daegwallyeong.
    The starting point \textit{T} is 2022-09-05 15:00 (UTC). 
    }
    \label{fig:limit_dae}
    \vspace{-3mm}
\end{figure}
\subsection{Limitations and Discussions}
\label{subsec:limitations}
PCT-CycleGAN does iterative forecasting, so it forecasts an $(N+1)$-step future from an $N$-step future.
Therefore, errors in the $N$-step future can lead to persistent errors at $(N+1)$-step future and beyond.
The failure case can be seen in~\Cref{fig:limit_seoul}.
The purple boxes show the mispredicted radar echoes.
The misprediction at $T+40$ min causes persistent errors.
It is an aspect that cannot be solved by qualitative or quantitative evaluation alone.
A supervisor algorithm or a more sophisticated loss function will be required.

The typhoon case nowcasting results are shown in~\Cref{fig:limit_dae}.
In this case, the prediction details of PCT-CycleGAN are not good.
In particular, empty-echo areas (black boxes) predicted by PCT-CycleGAN to be free of rain are hopeless.
This indicates that PCT-CycleGAN should learn more cases about unusual meteorological phenomena to be a perfect model.

\section{Conclusion}
\label{sec:conclusion}
We proposed a novel PCT-CycleGAN for precipitation nowcasting.
Two generator networks and two discriminator networks were trained in the paired complementary temporal cycles.
One of the generator networks learned forward temporal dynamics and the other learned backward temporal dynamics.
We proposed connection loss to make the temporal causality of PCT-CycleGAN more robust.
We also proposed torrential loss to improve performance in exceptional heavy rain events.
As a result, PCT-CycleGAN showed comparable performance (better in many cases) against the latest precipitation nowcasting methods without any other complex architectures or training skills.
This was demonstrated through evaluation using CSI, PSNR, and SSIM.

\bibliographystyle{ACM-Reference-Format}
\bibliography{pct-cyclegan}
\end{document}